IFoodCloud: A Platform for Real-time Sentiment Analysis of Public Opinion about Food Safety in China

Dachuan Zhang [b, †], Haoyang Zhang [c, †], Zhisheng Wei [d], Yan Li [c], Zhiheng Mao [c], Chunmeng He [c], Haorui Ma [c], Xin Zeng [c], Xiaoling Xie [c], Xingran Kou [e] and Bingwen Zhang [a, *]

[a] Department of Food Science and Nutrition, University of Jinan, No. 13, Shungeng Road, Jinan, 250002, P. R. China

[b] CAS Key Laboratory of Computational Biology, Shanghai Institute of Nutrition and Health, University of Chinese Academy of Sciences, Chinese Academy of Sciences, Shanghai, 200333, P. R. China

[c] College of Food Engineering and Nutritional Science, Shaanxi Normal University, Xi'an 710119, P. R. China

[d] College of Food, Shihezi University, Shihezi, 832003, P. R. China

[e] School of Perfume and Aroma, Shanghai Institute of Technology, Shanghai, 200333, P. R. China

† Contributed equally
* Corresponding author. E-mail address: st_zhangbw@ujn.edu.cn

**Abstract**

The Internet contains a wealth of public opinion on food safety, including views on food adulteration, food-borne diseases, agricultural pollution, irregular food distribution, and food production issues. In order to systematically collect and analyse public opinion on food safety, we developed IFoodCloud, a platform for the real-time sentiment analysis of public opinion on food safety in China. It collects data from more than 3,100 public sources that can be used to explore public opinion trends, public sentiment, and regional attention differences of food safety incidents. At the same time, we constructed a sentiment classification model using multiple lexicon-based and deep learning-based algorithms integrated with IFoodCloud that provide an unprecedented rapid means of understanding the public sentiment toward specific food safety incidents. Our best model's F1-score achieved 0.9737. Further, three real-world cases are presented to demonstrate the application and robustness. IFoodCloud could be considered a valuable tool for promote scientisation of food safety supervision and risk communication.

*Keywords:* Food safety; Database; Sentiment analysis; Deep learning; Risk communication

# 1. Introduction

In recent years, consumers have become alert to frequent food safety incidents. Meanwhile, rapid Internet development has allowed the public to express interests and vent about food safety incidents online. The enthusiasm with which people participate in public affairs has risen, and more people are discussing food safety incidents online.

In this era of "Big Data", scientific research has increasingly mined valuable data from large online public opinion sources (Marvin, Janssen, Bouzembrak, Hendriksen, & Staats, 2017; Zhou, Zhang, Liu, Qiu, & He, 2019). *Internet public opinion* is a set of emotions, attitudes, and opinions that are disseminated and interacted with, subsequently influencing the majority of netizens. Its carriers mainly include electronic newspapers, news websites, forums, Bulletin Board Systems (BBS), and social media, among others. Its defining characteristics are large data samples, high diffusivity, and timeliness. Many food safety issues, such as food adulteration, food-borne diseases, and agricultural pollution are reflected in public opinion (Bouzembrak, Steen, Neslo, Linge, Mojtahed, & Marvin, 2018; P. Liu & Ma, 2016; Liu, Liu, Zhang, & Gao, 2015). At the same time, the Internet is also the primary information channel used by the Chinese public to obtain food safety information (R. Liu, Pieniak, & Verbeke, 2014). Biased food safety information or fabricated food safety rumours transmitted online have the potential to increase the frequency of food scares (R. Liu, Pieniak, & Verbeke, 2013; Stevens, Aarts, Termeer, & Dewulf, 2018). Therefore, the systematic collection and analysis of online food safety public opinion is essential for food safety risk management and communication.

Previous research on public attitudes toward food safety issues has primarily relied on questionnaires or face-to-face interviews (Berhane et al., 2018; Massaglia, Merlino, Borra, Bargetto, Sottile, & Peano, 2019; Ritter, Shriver, McConnachie, Robbins, von Keyserlingk, & Weary, 2019; Xu, Wu, & Luan, 2020). These methods complicate data acquisition and result in relatively small sample sizes. Based on the rich and valuable information contained in the data on public food safety opinion, Liu et al. (2015) analysed 295 representative food safety incidents that occurred in China since 2001 and were included in "Throwing out the window net", which is a food safety warning website set up by Internet volunteers. Qiang, Wen, Jing, and Yue (2011) collected food safety reports from 43 food-related websites and conducted a content analysis. They investigated several subjects, such as the food categories, violative testing items, and hazard of violative food samples in food safety reports between April 1 and June 1, 2009. Chen, Huang, Nong, and Kwan (2016) obtained food safety-related information from more than 100 websites to construct a food safety information database for Greater China, comparing the differences between food-borne disease data as reported in governmental documents and as reported in media reports. Zhu, Huang, and Manning (2019) explored the media's role in China's dairy product safety management. They found that the Chinese government performed better on exposing incidents earlier than the media, but the media played a complementary role in food safety governance by providing wider incident coverage than the government. These studies have important significance for food safety risk communication. However, the scope of data acquisition in the most recent research is still insufficient and lacks timeliness, complicating the real-time study of major food safety incidents. Additionally, traditional survey methods face difficulties in providing objective evaluations of the collective public sentiment about specific food safety issues. Internet public opinion often contains the public's view regarding food safety issues, which can be analysed to capture the public sentiment. This method is called *sentiment analysis*, which is a rising research area in the natural language processing field. However, sentiment analysis is domain-specific. When the training set and test set belong to different fields, sentiment classification based on supervised learning usually shows poor effect. Therefore, the analysis models in other fields usually cannot achieve satisfactory results when analysing the public sentiment on food safety.

In this study, we constructed IFoodCloud, a platform for the real-time sentiment analysis of public opinion on food safety that also serves as a comprehensive database of public opinion. IFoodCloud automatically obtains data from over 3,100 public data sources and provides a variety of visual analysis tools for exploring attention-degree differences across regions and public opinion trends for specific food safety incidents. This provides powerful data support for scientific analysis, food safety supervision, and risk communication. We also constructed a sentiment classification model through multiple lexicon-based and deep-learning-based algorithms that can rapidly identify the future sentiment tendencies in the already dynamic public opinion on food safety. Further, three real-world cases are presented to demonstrate IFoodCloud's application and robustness. First, we used IFoodCloud data to compare differences in food safety public sentiment across mass media, social media, and government media. Second, we used IFoodCloud to construct a China-wide landscape of public opinion about domestic food safety that graphically shows food safety sentiment differences across regions, reflecting both the governance ability of local food safety regulatory agencies and public satisfaction with the regional food safety situation. Third, we used IFoodCloud to observe the changing trend of public opinion sentiment in China at the early stage of the Coronavirus Disease 2019 (COVID-19) pandemic to provide reference for government supervision.

**2. Materials and methods**

2.1. Data collection, storage, and management

Figure 1 shows the execution pipeline of IFoodCloud. The Internet data collection program is based on the use of Python's Scrapy, Selenium, and Requests for web scraping. This program obtains information on topics related to food safety from 2494 news websites, 466 electronic newspapers, and 238 BBS, forums, and blogs in China. We categorised the different news types according to their data sources. Government websites were categorised as *government media*, such as the China Food and Drug Administration website, representing government agencies' views on food safety status in China. We categorised information published by electronic newspapers and news websites as *mass media*, representing the views of the media industry and public figures on food safety status in China. Information from Weibo, forums, and BBS was categorised as *social media*, representing the general public's views and emotions on food safety. In order to ensure timeliness, the data acquisition program runs 24 hours a day. After the data is collected, the system performs data cleaning to remove invalid data. We also constructed a dictionary-based geographic location extractor based on province, city, and country name, as collected from the National Bureau of Statistics of China, which is used to extract geographic location information in the public opinion text. At the same time, the TextRank (Mallick, Das, Dutta, Das, & Sarkar, 2019) algorithm is used to extract abstracts and keywords from public opinion text so scientists can quickly browse the content of public opinion.

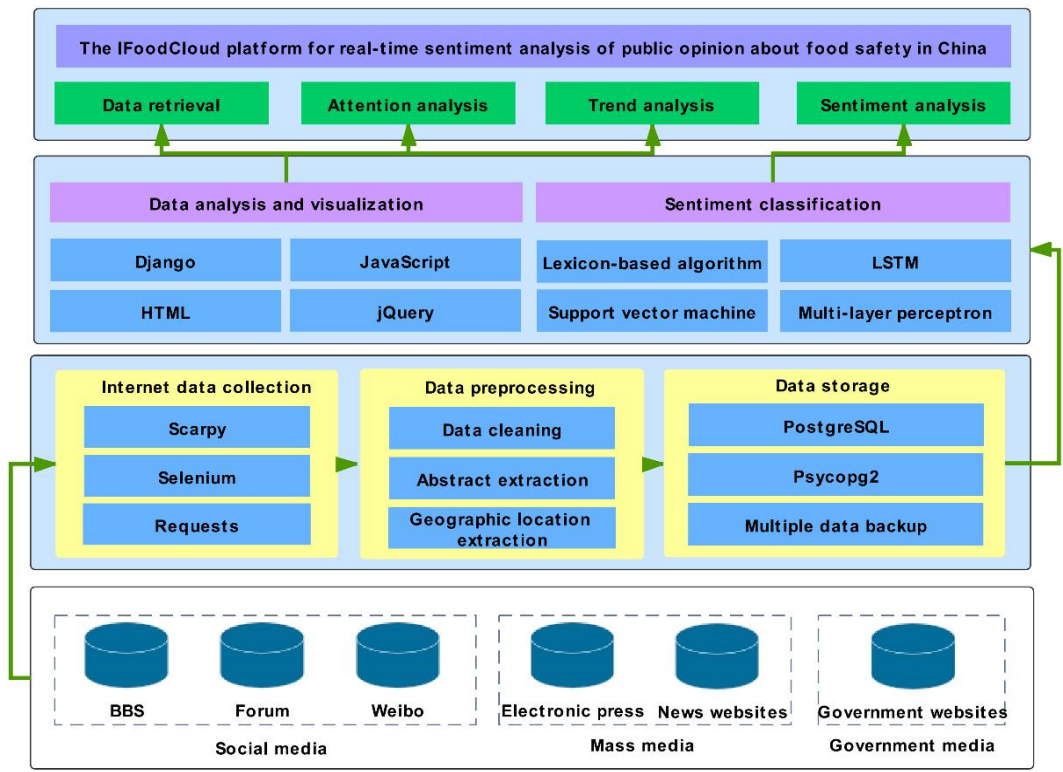

Figure 1. The framework of IFoodCloud includes the tools and algorithms used. IFoodCloud automatically performs data collection, cleaning, and storage. It runs on the Ubuntu Server system, stores data in a PostgreSQL database, builds a back-end framework based on Django, and implements front-end display and data visualization based on JavaScript, Bootstrap, HTML, and jQuery.

2.2. Food safety public opinion sentiment classification

2.2.1 Lexicon-based sentiment classification

*Lexicon-based sentiment classification* is a classification method based on linguistics that uses a series of dictionaries of pre-marked words, including a sentiment dictionary, degree dictionary, inversion dictionary, and stopword dictionary. A *sentiment dictionary* is a collection of sentiment words, and its domain specificity is critical to the accuracy of the model. We integrated specific food safety vocabulary and authoritative sentiment dictionaries. Finally, we obtained 33,202 sentiment words that included 14,476 positive words and 18,726 negative words. A *degree dictionary* is a collection of degree adverbs that represent the strength of a sentiment. The emotional strength of five subcategories, *most*, *very*, *more*, *nearly*, and *barely*, decreases in order. The adverbs assign weight to the corresponding sentiment words to calculate the sentiment score. An *inversion dictionary* is a collection of inversion words. The inversion dictionary is used to determine whether to reverse the sentiment of the vocabulary. A *stopword dictionary* is a collection of predefined meaningless words, usually articles, prepositions, adverbs, or conjunctions (i.e., *the*, *is*, *at*, *which*, and *on*) that are used to remove meaningless words from the public opinion text.

The lexicon-based sentiment classification of public opinion follows four steps (Fig. 2). First, public opinion text is divided into multiple words though Python's Jieba package, then meaningless words are deleted based on the stopword dictionary. Second, sentiment words are identified using the predefined sentiment dictionary. Based on a sentiment word's position in a sentence, our program draws from the other two dictionaries to conduct automatic searches for degree adverbs and inversion words that modify the sentiment word. Third, the sentiment score of each sentiment word is calculated based on the weight of its degree adverbs and inversion words. Fourth, the sentiment score of public opinion text is calculated. The scores of all sentiment words in the public opinion text are then added to obtain the final sentiment score. When sentiment score is positive, the text sentiment is positive; otherwise, the text sentiment is negative. Meanwhile, a higher sentiment score indicates that stronger sentiment. This method is simple, easy to operate, and can simultaneously obtain both the sentiment tendency and sentiment intensity of public opinion.

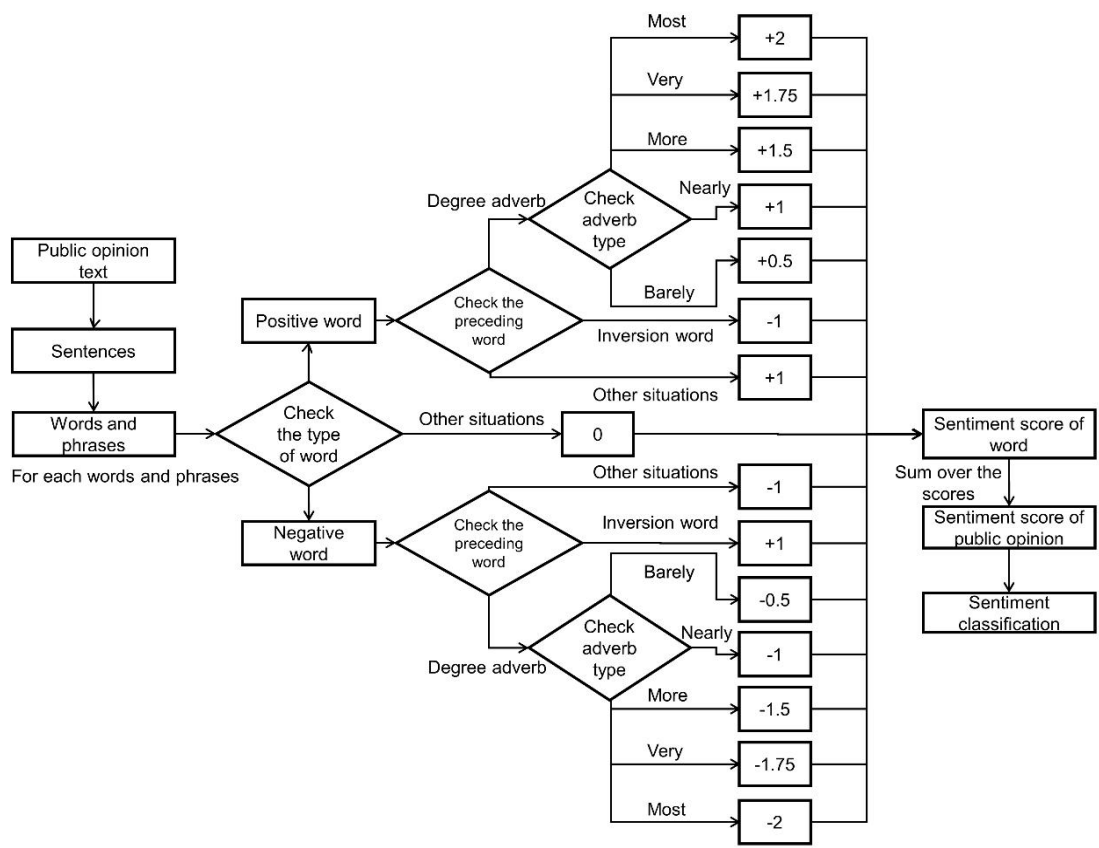

Figure 2. The lexicon-based sentiment classification pipeline. The public opinion text was divided into multiple sentences, then further divided into multiple words and phrases. Next, each word and phrase is checked. If a word is a sentiment word, then its modifying adverbs are identified to determine its sentiment intensity, according to the type of adverb. Finally, the sentiment scores of all words and phrases are summed to obtain the overall sentiment score of the public opinion text, and the sentiment tendency is judged based on the score.

2.2.2 Deep learning-based sentiment classification

In addition to the lexicon-based method, we also used state-of-the-art deep learning methods to construct sentiment classification models. *Deep learning* is a computer science that allows computers to learn from data. It mainly consists of four phases: (1) acquiring data, (2) data pre-processing, (3) training models and (4) evaluating model. We manually labelled the sentiment of partial public opinion text in IFoodCloud and randomly selected 7,500 positive and 2,500 negative public opinion texts to form the training and test model dataset. After that, Word2Vec (Mikolov, Chen, Corrado, & Dean, 2013), a technology that extracts useful information from a large amount of text through neural networks, was used to pre-process the data by converting the text into word vectors containing semantic information. This technology solved the data sparseness and dimensional disasters problems that often occur in traditional vectorization methods and has been widely used to explore knowledge from a large number of texts (Tshitoyan et al., 2019). We then used two deep learning algorithms to train the model: Multi-Layer Perceptron (MLP) and Long Short-Term Memory (LSTM). MLP is an artificial neural network with a forward structure that maps a group of input vectors to a group of output vectors. MLP can be thought of as a directed graph, consisting of multiple node layers that are each fully connected to the next layer. With the exception of the input nodes, each node is a neuron with a nonlinear activation function. LSTM is a form of Recurrent Neural Network (RNN) implementation, which has memory timing characteristics and can learn the correlation between data contexts. The LSTM model introduces memory units and door mechanisms to process historical information, which mainly covers the input, output, and forget gates. The LSTM mechanism for processing historical information is as follows:

$$i_t = \sigma(W_{xi}x_t + W_{hi}h_{t-1} + W_{ci}c_{t-1} + b_i) (Eq.1)$$

$$f_t = \sigma(W_{xf}x_t + W_{hf}h_{t-1} + W_{cf}c_{t-1} + b_f) (Eq.2)$$

$$c_t = f_t \odot C_{t-1} + i_t \odot tanh(W_{xi}x_t + W_{hi}h_{i-1} + b_i) (Eq.3)$$

$$O_t = \sigma(W_{xo}x_t + W_{ho}h_{t-1} + W_{co}c_{t-1} + b_o) (Eq.4)$$

$$h_t = O_t \odot \tanh(c_t) \ (Eq.5)$$

In Eq. 1-5, *W* and *b* are model parameters, and *tanh* is a hyperbolic tangent curve. $i_t$, $f_t$, $o_t$ and $c_t$ respectively represent the input gate, forget gate, output gate, and memory cell output at time *t* of food safety public opinion sentiment classification model training (Fig.3). For comparison, we also tested the performance of the classical machine learning model, Support Vector Machine (SVM), and one of the best general Chinese sentiment classification models, Senta, on our dataset.

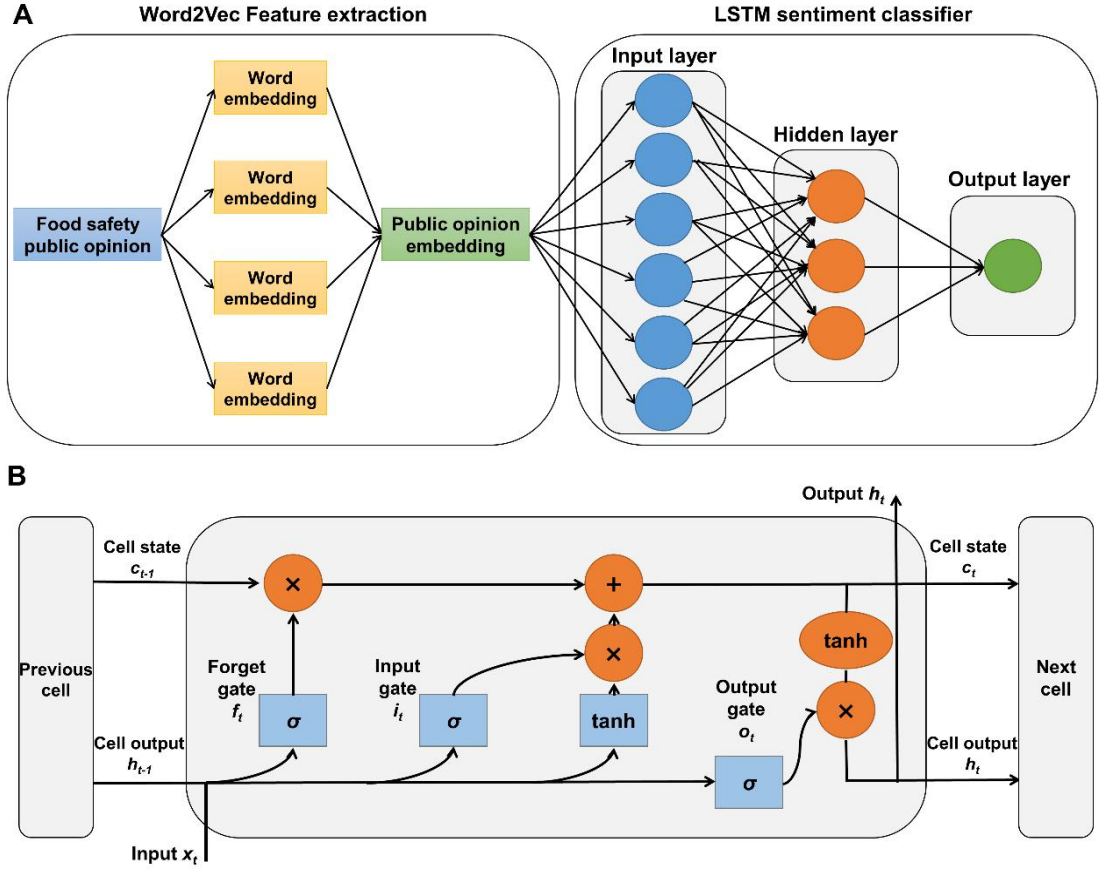

Figure 3. Structure of the long short-term memory (LSTM) model. (A) The architecture of the LSTM sentiment classification model. (B) The information flow and components of an LSTM cell.

2.2.3 Model evaluation

We used accuracy, recall, and F1-score to evaluate the model's classification performance. *Precision* refers the predicted correct proportion of an instance with a positive sample (Eq.6). *Recall* refers to the ratio of the number of correctly predicted positive samples to the total number of true positive samples (Eq.7). *F1-score* is the harmonic average of precision and recall, comprehensively referring to these two indicators (Eq.8).

$$Precision = \frac{TP}{TP + FP} \ (Eq.6)$$

$$Recall = \frac{TP}{TP + FN} \ (Eq.7)$$

$$F1 - score = \frac{2 \times TP}{2 \times TP + FP + FN} \ (Eq.8)$$

*TP, TN, FP,* and *FN* are the numbers of true positive public opinion, true negative public opinion, false positive public opinion, and false negative public opinion, respectively.

2.3 Analysis and visualization of public sentiment for specific food safety incidents or regions

We used the *positive public opinion ratio (PPR)* to reflect the public sentiment toward specific food safety incidents or regions. When researching a specific food safety incident, we matched all relevant public opinion to the time series, then calculated the public opinion number and PPR of each day in the time series to illuminate the attention trends and public sentiment changes. Only when the public opinion on a given day reached a certain sample size (frequency greater than 10) was the sentiment counted. Otherwise, the public opinion on the given day was considered to indicate a non-sentiment tendency (PPR=0.5), due to lack of sufficient data support. When analysing relative attention, we used the relative attention index to report the level of public awareness of specific food safety incidents in different regions. This was reflected by the number of public opinion texts on specific food safety incidents in a region as a proportion of the number of all relevant food safety public opinions. Finally, ECharts (https://echarts.apache.org/) was used to visualize the calculated data.

## 3. Results and discussion

3.1. Platform overview

Since its launch in June 2019, IFoodCloud has obtained more than 100,000 pieces of food safety public opinion data from more than 3,100 public data sources, containing 75,051 mass media, 21,684 social media, and 4,298 government media observations. For each of these observations, our data contain the public opinion title, content, publication time, data source, Uniform Resource Locator, extracted abstract, extracted keywords, extracted geographic location information, predicted sentiment category, and predicted sentiment score. Compared with previous studies, the scope and quantity of our study's data acquisition mark significant improvement. IFoodCloud provides search methods, such as keyword and time range searches (Fig. 4). After the search is completed, IFoodCloud provides further analysis functions, including food safety incident trend, sentiment, and regional attention difference analyses.

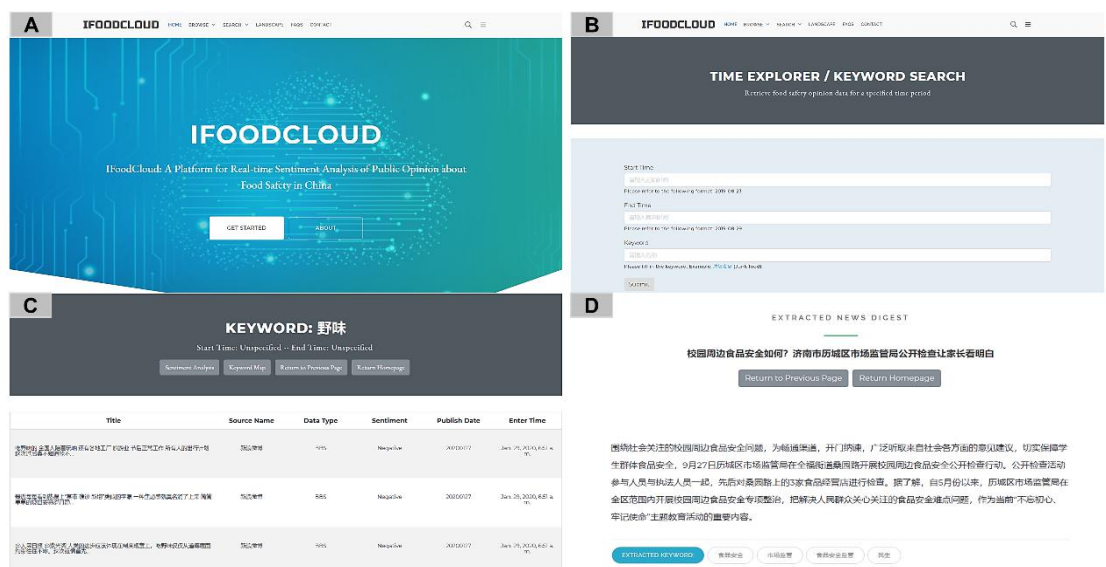

Figure 4. IFoodCloud interface. (A) Home page. (B) Search interface. Keywords and time range can be used as search criteria. (C) Public opinion data retrieved with "野味 (Exotic animals)" as the keyword. (D) Public opinion details page.

3.2. Sentiment classification model of food safety public opinion

Table 1 shows the precision, recall, and F1-score of multiple models. The general sentiment classification model, Senta, achieved high precision at 0.9648, but the recall and F1-score are relatively low, at 0.3308 and 0.4927, respectively. In other words, Senta judged many positive public opinion instances as negative sentiment. This reflects the general model's difficulty in understanding the emotions of food safety-related texts and underscores the necessity of constructing a special sentiment classification model for food safety public opinion.

Table 1. The performance of each model in sentiment classification.

| Model | Precision | Recall | F1-score |
| --- | --- | --- | --- |
| Senta | 0.9648 | 0.3308 | 0.4927 |
| Lexicon-based model | 0.9025 | 0.9508 | 0.9260 |
| SVM | 0.9628 | 0.9628 | 0.9697 |
| MLP | 0.9594 | 0.9690 | 0.9641 |
| LSTM | 0.9720 | 0.9754 | 0.9737 |

The lexicon-based model's precision, recall, and F1-score were 0.9025, 0.9508, and 0.9260, respectively. Although the model implementation is relatively simple, it still performed much better than general sentiment model. Additionally, it was able to gather sentiment intensity at the same time. The SVM model's precision, recall, and F1-score were 0.9628, 0.9628 and 0.9697, respectively. The MLP model's precision, recall, and F1-score were 0.9594, 0.9690 and 0.9641, respectively. Both models achieved good performance, but the SVM model's performance was slightly better than that of the MLP model. This is potentially because the deep learning model requires a high number of training samples (million-plus), making it difficult to achieve better performance with relatively small training samples. The LSTM model had the best performance, with a precision, recall, and F1-score of 0.9720, 0.9754, and 0.9737, respectively. This is likely because LSTM introduces cellular state and gate state for state transfer and control, which can better solve the problem of long-term dependence in sequence. In natural language processing, it also helps to retain the semantic information of the whole sequence and eliminate useless state information. This solves the problem that deep learning models can face when presented with relatively long sentence input, as the models can "forget" some relevant information that is too far from the focus words, even when the sentence's meaning is dependent on the "forgotten" words.

3.3 Application cases

3.3.1 Contrast the sentiment differences between different types of media

We used the best-performing LSTM model to analyse the sentiment expressed in the food safety opinions collected in IFoodCloud from June 2019 to March 2020. We then calculated the PPR.

Meanwhile, the lexicon-based model was used to calculate the sentiment score.

The results showed an overall PPR of public opinion of 0.8693, indicating that the overall public opinion of food safety in China is relatively positive. Among the different media types, the mass media PPR was 0.9158, the government media PPR was 0.9581, and the social media PPR was 0.6911. The government media's public opinion had the highest PPR, indicating that the government is more inclined to express positive sentiment to the public to avoid causing public panic over food safety incidents. Figure 5 shows the government media's wider distribution of positive sentiments alongside the generally lower, more concentrated negative sentiment scores. The distribution of mass media's sentiment scores is similar to the government media distribution, but it ranges further into very low sentiment scores (Fig. 5A.B). This reflects China's primarily positive mass media coverage, though mass media also harshly criticizes some odious food safety incidents. The social media PPR was relatively lower, and both positive and negative sentiment scores were more widely distributed and more concentrated at zero (tending to be neutral). A high proportion of negative opinions and a low sentiment intensity indicate that social media may objectively reveal more numerous food safety issues that concern the public with a voice of reason, making social media more valuable than mass media as a decision-making reference for management agencies.

3.3.2 China-wide landscape of public sentiment about domestic food safety

Based on public opinion and the related geographical location information recorded in IFoodCloud from June 2019 to Jan 2020, we matched all the public opinion data to China's 34 administrative divisions and calculated the relevant public opinion PPR in each administrative division. Further, we used the above results and the ECharts visual tool to construct a China-wide landscape of public sentiment toward domestic food safety (Fig. 5C).

Public food safety opinion in Henan (PPR: 0.9239), Inner Mongolia (PPR: 0.9142), Shandong (PPR: 0.9122), and Shanxi (PPR: 0.9086) was more positive. Public food safety opinion in Ningxia (PPR: 0.7183), Tibet (PPR: 0.7477), Chongqing (PPR: 0.7681), Liaoning (PPR: 0.7686), and Hubei (PPR: 0.7738) was relatively negative. There was a significant difference in the public sentiment toward food safety between the provinces. Results show that the public sentiment in northern areas was generally higher than in southern areas, and the public sentiment of neighbouring provinces was often close. This is potentially because food commerce and trade is more frequent between closer provinces, so the overall food safety situation is similar. Further, we found that the public sentiment in provinces with less-developed economies, such as Ningxia, Tibet, Hainan, and Qinghai, were lower. According to data from the National Bureau of Statistics of China and the World Bank, the 2018 GDP of these provinces ranked 32nd, 34th, 30th, and 33rd, respectively, among the 34 administrative divisions in China. Public sentiment about food safety is affected by multiple factors, so the relationships among public sentiment about food safety in different regions and local economic, culture, public health, and social conditions are worthy of further study.

It is worth noting that Hubei Province has a relatively developed economy (7th of China's 34 administrative divisions), and the overall public sentiment in surrounding provinces is relatively high.

However, the province's own public sentiment is relatively low. This is because of the outbreak of COVID-19 in Hubei, which affected the province's overall public opinion sentiment. The possible cause of the epidemic is the consumption of wild animals (Calisher, Carroll, Colwell, Corley, Daszak, Drosten, et al., 2020) such as bats, paguma larvata, and pangolins. This was reflected in the public mentions of food safety when discussing the epidemic, even though this is not a univocal food safety incident. This public opinion was detected by IFoodCloud. We tested the PPR public opinion related to COVID-19 and the PPR of public opinion unrelated to COVID-19. The former was 0.4392, and the latter was 0.8648. The above results show that major food safety and derivative incidents can lead to a significant decline in public sentiment in relevant regions.

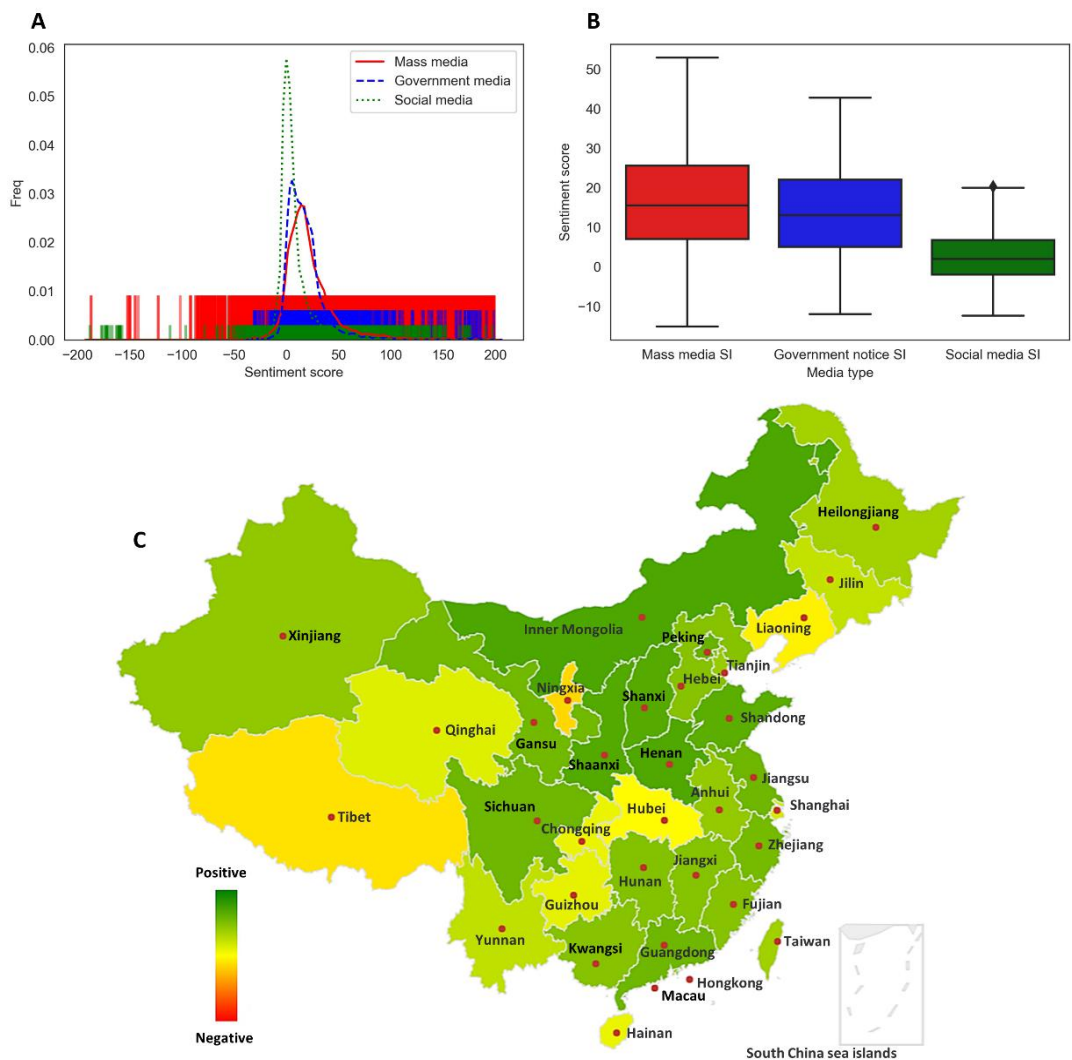

Figure 5. (A) Histogram of the public sentiments toward mass media, social media, and government media on food safety. (B) Box plot of public sentiments toward mass media, social media, and government media on food safety. (C) Global landscape of food safety public sentiment across China,

ranging from negative emotions, represented by red, to positive emotions, represented by green. The red dots mark the provincial capitals.

3.3.3 Auxiliary government supervision based on the changing trend of public opinion sentiment

The COVID-19 outbreak has significantly affected global health and society. As of May 11, 2020, the incident had infected 4,114,379 people and killed 279,626 people worldwide. Taking the early stage of the COVID-19 outbreak as an example, we selected the public opinion data relevant to COVID-19 from January 1, 2020 to January 31, 2020 in the IFoodCloud database, analysing attention change trends, sentiment change trends, and regional attention differences.

As shown in Figure 6A, prior to January 20, the number of relevant public opinions was low, and the public sentiment was stable. This indicates that, at that time, the incident had not yet attracted public attention. On January 20, there was a slight increase in the number of relevant public opinions, with a PPR of 0.8571, indicating that the incident had begun to attract public attention. Moreover, the high public sentiment indicates that the public had confidence in the government to resolve the incident. Beginning on January 22, the public's attention to the incident began to increase rapidly. At the same time, the public sentiment began to decline. Public sentiment reached a low on January 23 (PPR: 0.4928), indicating that, at that time, the general public was panicking about the incident. The likely reason for this decline in public sentiment is that Wuhan City had been completely sealed off to control the spread of COVID-19, conveying the incident's gravity to citizens. After January 23, public attention toward the incident began to decline, and the related public opinion grew more positive. Beginning January 27, public attention toward COVID-19 began to increase rapidly, and public sentiment continued to decline, reaching a low on January 28 (PPR: 0.4933). From the perspective of public attention in each province, Hubei had the highest level of public attention (Figure 6B). Overall, public attention in each province decreased as geographical distance from Hubei increased, except in Guangdong, Sichuan, Shanxi and Jiangsu, all of which attracted greater public attention than provinces closer to Hubei.

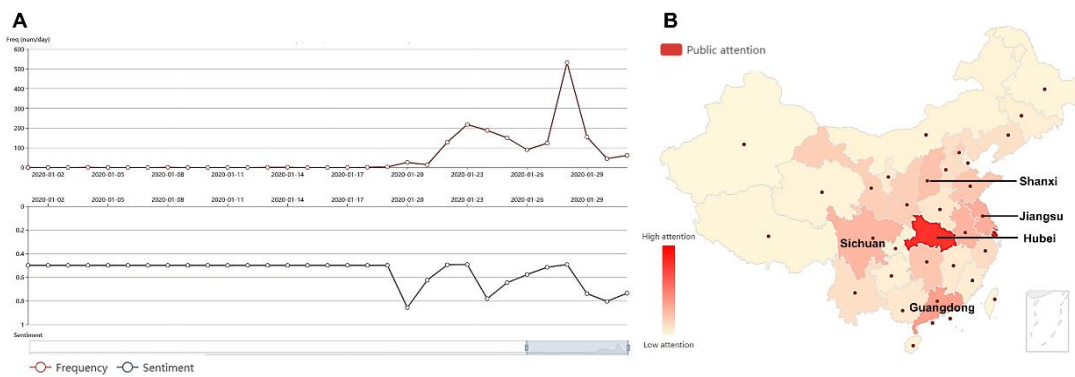

Figure 6. (A) Attention and sentiment changes in the 2020 coronavirus incident. (B) Relative attention of provinces in China to this incident.

The above cases clearly reflect that the government should have actively intervened on January 22, 2020 when the growth curve of public attention and sentiment showed a steeper trend. Authorities should give timely explanations to the public by considering the content of online public opinion in IFoodCloud to maintain social stability and enhance public confidence in the management agency's governance capabilities. The provinces with a higher degree of relative concern, such as Hubei, and provinces with more relative concern than surrounding regions, such as Guangdong, Sichuan, Shanxi and Jiangsu, need to strengthen pandemic prevention and media guidance more than others.

When other food safety-related incidents occur, managers, authorities, scientists, and other stakeholders can quickly gather and study the public sentiment toward these incidents through IFoodCloud and explore the incidents' impact across different regions. Thus, governments and scientists can effectively manage food safety incidents and carry out related food safety risk communication in a timely manner.

## 4. Conclusion

IFoodCloud is a platform for the real-time sentiment analysis of public opinion about food safety in China, as well as a comprehensive database that provides high-quality corpus data and has the potential to become a benchmark for future food safety information mining. Compared with traditional survey methods, using data from the Internet public opinion about food safety allows for larger sample sizes, more comprehensive data coverage, lower data collection and analysis costs, and more timely results. It provides a meaningful supplement to actual investigation into food safety issues and also serves as a valuable reference for food safety emergency response. Meanwhile, the sentiment classification model that we built after testing the performance of multiple algorithms can be used to quickly identify the future sentiment tendencies in an already dynamic food safety public opinion.

Our three application cases found that online public opinion about food safety in China is generally positive. Mass media has the highest level of positive sentiment, followed by government media. Social media has the lowest level of positive sentiment. We also found significant differences in public sentiment regarding food safety opinions across different regions of China. Such differences in public sentiment have the potential to become a reference for evaluating the governing ability of local food safety regulatory agencies and the public's satisfaction with the regional food safety situation.

In addition to the application cases shown above, the platform proposed in this study is also suitable to help food developers and manufacturers discover market trends from public opinion data and evaluate products' market performance. In our future research, we will expand the data acquisition scope of IFoodCloud. At the same time, we plan to develop novel analysis functions on the basis of the data system constructed in this work, such as incident correlation analysis and public opinion theme analysis. We also plan to develop a specific model to predict public opinion development trends and provide early warnings. Based on these potential applications, we expect that IFoodCloud will become an indispensable tool for use in future public opinion studies on food safety.

## 5. Declaration of interest

Declarations of interest: none

## 6. Funding and acknowledgements

This work was supported by the National Social Science Foundation of China [grant number 13BGL096], and the Soft Science Plan Project of Jinan [grant number 201502151].